\documentclass{sip}

\usepackage{amsmath,color,bm,amssymb,mathrsfs,subfigure,booktabs,url,mathrsfs}
\usepackage{graphicx}

\definecolor{amazonite}{RGB}{0,115,150}
\definecolor{myred}{RGB}{255,56,0}
\definecolor{momoiro}{RGB}{240,145,153}
\definecolor{mygreen}{RGB}{30,150,30}
\definecolor{mybrown}{RGB}{150,30,30}
\definecolor{aomurasakiiro}{RGB}{103,69,152}

\begin{document}

\title[Statistical Piano Reduction]{
Statistical Piano Reduction Controlling Performance Difficulty
}

\author[Eita Nakamura, \textit{et al}.]{Eita Nakamura$^{1}$ and Kazuyoshi Yoshii$^{1,2}$}

\address{\add{1}{Graduate School of Informatics, Kyoto University, Kyoto 606-8501, Japan}
\add{2}{RIKEN Center for Advanced Intelligence Project, Tokyo 103-0027, Japan}}

\corres{\name{Eita Nakamura}
\email{enakamura@sap.ist.i.kyoto-u.ac.jp}}

\begin{abstract}
We present a statistical-modelling method for piano reduction, i.e.\ converting an ensemble score into piano scores, that can control performance difficulty. While previous studies have focused on describing the condition for playable piano scores, it depends on player's skill and can change continuously with the tempo. We thus computationally quantify performance difficulty as well as musical fidelity to the original score, and formulate the problem as optimization of musical fidelity under constraints on difficulty values. First, performance difficulty measures are developed by means of probabilistic generative models for piano scores and the relation to the rate of performance errors is studied. Second, to describe musical fidelity, we construct a probabilistic model integrating a prior piano-score model and a model representing how ensemble scores are likely to be edited. An iterative optimization algorithm for piano reduction is developed based on statistical inference of the model. We confirm the effect of the iterative procedure; we find that subjective difficulty and musical fidelity monotonically increase with controlled difficulty values; and we show that incorporating sequential dependence of pitches and fingering motion in the piano-score model improves the quality of reduction scores in high-difficulty cases.
\end{abstract}


\maketitle

\section{Introduction}
\label{sec:Intro}

\begin{figure}[tb]
\centering
\includegraphics[clip,width=0.95\columnwidth]{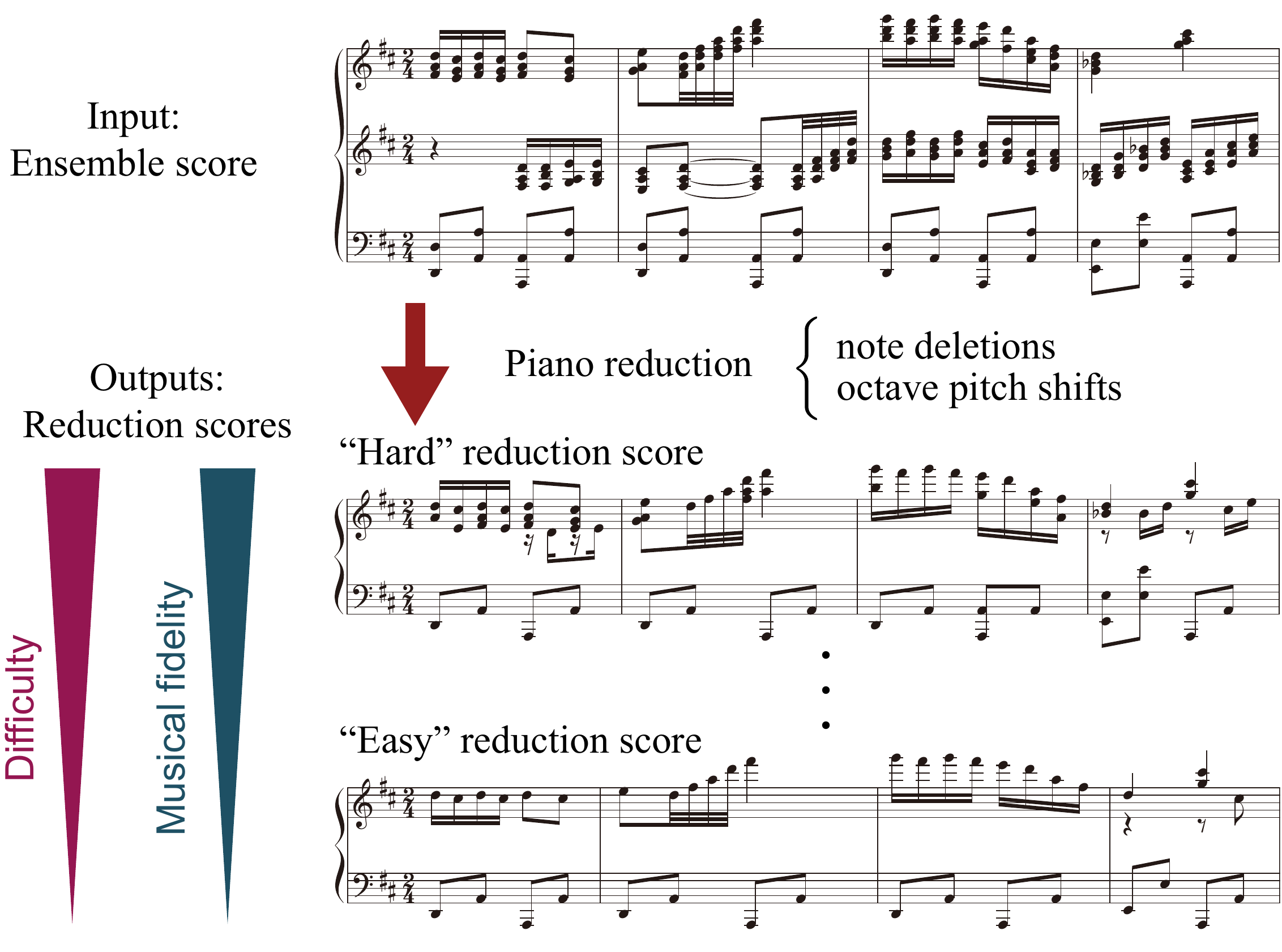}
\caption{Overview of the proposed system for piano reduction that can control performance difficulty.}
\label{fig:SystemOverview}
\vspace{-4mm}
\end{figure}

Music arrangement involving a change of instrumentation (e.g.\ arrangement for piano, guitar, etc.) is an important process of music creation to increase the variety of music performances.
Arranging a musical piece to change difficulty, for example, to make it playable for beginners, is also widely practiced.
To automate these processes, systems for piano arrangement \cite{Chiu2009,Onuma2010,Huang2012,Nakamura2015,Takamori2017}, guitar arrangement \cite{Tuohy2005,Hori2012,Hori2013}, and orchestration \cite{Maekawa2006,Crestel2017} have been studied.
This study aims at a system for piano reduction, i.e.\ converting an ensemble score (e.g.\ orchestral and band scores) into a piano score that can control performance difficulty and retain as much musical fidelity to the original score as possible (Fig.~\ref{fig:SystemOverview}).

To computationally judge whether a musical score is playable, previous studies have developed conditions on the pitch and rhythmic content.
For piano scores, conditions such as `there can be at most 5 simultaneous notes for each hand' and `simultaneous pitch spans for each hand must be less than 14 semitones (or so)' have been considered \cite{Onuma2010,Huang2012}.
However, these conditions cannot be thought of as necessary nor sufficient conditions for playable scores because in reality there can be a piano score with chords with more than 5 notes and/or spanning a large pitch interval that are conventionally played as broken chords, and even scores without chords (melodies) can be unplayable in fast tempos.
In fact, it is difficult to find a complete description of playable scores that is valid in every situation because the condition depends on player's skill and can change continuously with the tempo.
A possible solution is to quantify performance difficulty and use it as an indicator of playable scores in each situation of skill level, tempo, etc.\ \cite{Chiu2012,Sebastien2012}.

As there is generally a trade-off between performance difficulty and musical fidelity to the original score, it is necessary to quantify musical fidelity and develop an optimization method.
Music arrangers remove notes and shift pitches in an ensemble score for the piano reduction score to match a target difficulty level \cite{Onuma2010,Nakamura2015}.
From a statistical point of view, one can assign probabilities for these edit operations and use them to quantify musical fidelity.
Following the analogy with statistical machine translation \cite{StatisticalTranslation}, if one can construct a model for the probability $P(R|E)$ of a reduction score $R$ given an ensemble score $E$, the piano reduction problem can be formulated as optimization of $P(R|E)$ under constraints on difficulty values.
A similar approach without controls of performance difficulty has been studied for guitar arrangement \cite{Hori2012,Hori2013}.

To realize this idea, a statistical-modelling approach for piano reduction that can control performance difficulty has been proposed in a recent conference paper \cite{Nakamura2015}.
Following the thought that fingering motion is closely related to the cost or difficulty of performance \cite{Parncutt1997,Hart2000,Kasimi2007}, quantitative measures of performance difficulty were developed based on a probabilistic generative model of piano scores incorporating fingering motion \cite{Yonebayashi2007,Nakamura2014}.
To estimate the probability $P(R|E)$, a hidden Markov model (HMM) integrating the piano-score model and a model representing how ensemble scores are likely to be edited was constructed.
A piano reduction algorithm was developed based on the Viterbi algorithm.
While the potential of the method was suggested by the results of piano reduction for one example piece, formal evaluations and comparisons with other approaches were left for future work.
There was also a problem of the optimization method that the upper-bound constraints on difficulty values were often not properly satisfied, due to the limitation of the Viterbi algorithm.

In this study, we extend the work of \cite{Nakamura2015} and propose an improved piano reduction method using iterative optimization.
We also carry out systematic evaluations on the difficulty measure and the piano reduction method.
In particular, we evaluate difficulty measures in terms of their ability of predicting performance errors, which is to our knowledge the first attempt in the literature to objectively evaluating performance difficulty measures.
Piano reduction methods are evaluated both objectively and subjectively: an objective evaluation is conducted to examine the effect of the iterative optimization strategy; an subjective evaluation is conducted to assess the quality of the generated reduction scores.
The main results are:
\begin{itemize}
\item The proposed difficulty measures can be used as indicators of performance errors and measures incorporating the sequential nature of piano scores can better predict performance errors.
\item The proposed iterative optimization method yields better controls of difficulty than the method in \cite{Nakamura2015}.
\item Both subjective difficulty and musical fidelity of generated reduction scores monotonically increase with controlled difficulty values.
\item By comparing methods based on different models, it is shown that incorporating sequential dependence of pitches and fingering motion in the piano-score model improves musical naturalness and the rate of unplayable notes of reduction scores in high-difficulty cases.
\end{itemize}
The following are limitations of the current system:
\begin{itemize}
\item Melodic and bass notes are manually indicated.
\item Score typesetting, especially estimation of voices within each hand part, is currently done manually.
\end{itemize}
Automating these processes is an undeniable direction for future work.
See section \ref{sec:Evaluation}.\ref{sec:ExamplesAndDiscussions} for discussions.

The rest of the paper is organized as follows.
In the next section, we discuss generative piano-score models and performance difficulty measures.
In section \ref{sec:Method}, we present our method for piano reduction.
In section \ref{sec:Evaluation}, we present and discuss results of evaluation of the piano reduction method.
We conclude the paper in the last section.


\section{Quantitative Measures of Performance Difficulty}
\label{sec:FingeringModelAndDifficulty}

We formulate quantitative performance difficulty measures based on probabilistic generative models of piano scores.
A generative model incorporating piano fingering and simpler models are described in section \ref{sec:FingeringModelAndDifficulty}.\ref{sec:FingeringModel} and performance difficulty measures are discussed in section \ref{sec:FingeringModelAndDifficulty}.\ref{sec:Difficulty}.

\subsection{Generative Models for Piano Scores}
\label{sec:FingeringModel}

\subsubsection{Models for One Hand}
\label{sec:ModelForOneHand}

Let us first discuss models for one hand.
A piano score is represented as a sequence of pitches $p_{1:N}=(p_n)_{n=1}^N$ and corresponding onset times $t_{1:N}=(t_n)_{n=1}^N$ ($N$ is the number of musical notes).
A generative model for piano scores (piano-score model) is here defined as a model that yields the probability $P(p_{1:N})$.

Simple piano-score models can be constructed based on the Markov model.
The probability $P(p_{1:N})$ is factorized into an initial probability $P(p_1)$ and the transition probabilities $P(p_n|p_{n-1})$ as
\begin{equation}
P(p_{1:N})=P(p_1)\prod_{n=2}^NP(p_n|p_{n-1}).
\end{equation}
The simplest model is obtained by assuming that the initial and transition probabilities obey a uniform distribution over pitches.
Writing $N_p=88$ for the number of possible pitches, the model yields $P(p_{1:N})=(1/N_p)^N$.
Since this model yields the same probability for any piano score of the same length, it is here called a {\it no-information model}.

A more realistic model can be build by incorporating sequential dependence of pitches.
For example, a statistical tendency called pitch proximity, that successive pitches tend to be close to each other, can be incorporated in initial and transition probabilities described with Gaussians:
\begin{align}
P(p_1=p)&\propto {\rm Gauss}(p;p_0,\sigma_p^2)+\epsilon,
\\
P(p_n=p\,|\,p_{n-1}=p')&\propto {\rm Gauss}(p;p',\sigma_p^2)+\epsilon.
\end{align}
Here, ${\rm Gauss}(\,\cdot\,;\mu,\sigma^2)$ denotes a Gaussian distribution with mean $\mu$ and standard deviation $\sigma$, $p_0$ is a reference pitch to define the initial probability, and $\epsilon$ is a small positive constant for smoothing the probability for pitch transitions with a large leap.
We call this model a {\it Gaussian model}.

Although the Gaussian model can capture the tendency of pitch proximity, the simplification can lead to unrealistic consequences.
First, pitch transitions involving 10 or 11 semitones have higher probabilities than octave motions, which opposes the reality \cite{Nakamura2014}.
Second, since the model does not distinguish white keys and black keys, it yields the same probability for piano scores transposed to any keys, which opposes the fact that ``simpler keys'' involving less black keys are more frequently used.
In general, the difficulty or naturalness of a piano score changes when it is transposed to another key since the geometry of the piano keyboard requires different fingering motions \cite{Chiu2012}.
To solve this, it is necessary to construct a model that describes fingering motions in addition to pitch transitions.

\begin{figure}[tb]
\centering
\includegraphics[clip,width=0.97\columnwidth]{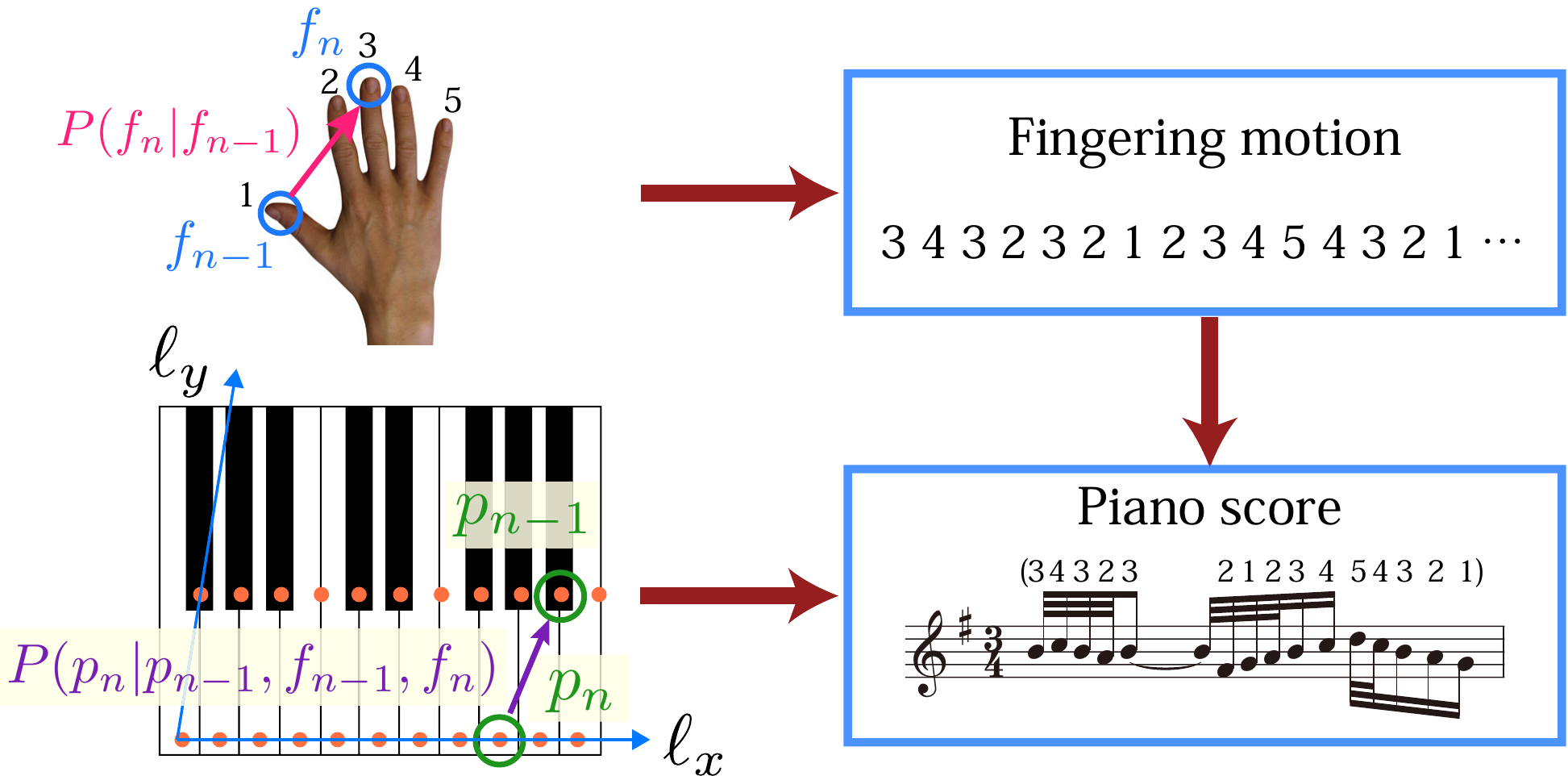}
\caption{Piano-score model incorporating fingering motion.}
\label{fig:FingeringModel}
\vspace{-2mm}
\end{figure}

A model (called {\it fingering model}) incorporating fingering motions and the geometry of the piano keyboard has been proposed in \cite{Nakamura2014}.
The model is based on HMM, which has been first applied to the piano fingering model in \cite{Yonebayashi2007}.
In general, we can introduce a stochastic variable $f_n$ representing a finger used to play the $n$\,th note.
The variable $f_n$ takes one of the following five values: $1=$ thumb, $2=$ index finger$,\cdots,$ $5=$ little finger \footnote{In this study, we do not consider the possibility of finger substitutions where two or more fingers are assigned to a note.}.
According to the model (Fig.~\ref{fig:FingeringModel}), a fingering motion $f_{1:N}=(f_n)_{n=1}^N$ is first generated by an initial probability $P(f_1)$ and transition probabilities $P(f_n|f_{n-1})$.
Next, a pitch sequence $p_{1:N}$ is generated conditionally on $f_{1:N}$: the first pitch is generated by $P(p_1|f_1)$ and the succeeding pitches are generated by $P(p_n|p_{n-1},f_{n-1},f_n)$, which describes the probability that a pitch would appear following the previous pitch and the previous and current fingers.
Thus, the joint probability of pitches and fingering motion $P(p_{1:N},f_{1:N})$ is given as
\begin{align}
P(f_1)P(p_1|f_1)\prod_{n=2}^NP(f_n|f_{n-1})P(p_n|p_{n-1},f_{n-1},f_n).
\notag
\end{align}
In general, the parameters of the fingering model can be learned from music data with pitches and annotated fingerings.
For want of a sufficient amount of data, the probability $P(p_n|p_{n-1},f_{n-1},f_n)$, which have $88^2\cdot 5^2$ parameters, cannot be trained effectively in a direct way.
We thus introduce simplifying assumptions to reduce the number of parameters.
First, we assume that the probability depends on pitches through their geometrical positions on the keyboard (Fig.~\ref{fig:FingeringModel}).
The coordinate on the keyboard of a pitch $p$ is represented as $\bm\ell(p)=(\ell_x(p),\ell_y(p))$.
We also assume translational symmetry in the $x$-direction and time inversion symmetry, which is expressed as
\begin{align}
&P(p_n=p\,|\,p_{n-1}=p',f_{n-1}=f',f_{n}=f')
\notag
\\
&\quad=F(\ell_x(p)-\ell_x(p'),\ell_y(p)-\ell_y(p');f',f)
\notag
\\
&\quad=F(\ell_x(p')-\ell_x(p),\ell_y(p')-\ell_y(p);f,f').
\end{align}
We also assume reflection symmetry between left and right hands.
The above model can be extended to including chords, by sequencing the contained notes from low pitch to high pitch \cite{Kasimi2007}.
With the fingering model, one can estimate the fingering $f_{1:N}$ from a given sequence of pitches $p_{1:N}$ by calculating the maximum of the probability $P(f_{1:N}|p_{1:N})\propto P(p_{1:N},f_{1:N})$.
This maximization can be computed by the Viterbi algorithm \cite{Rabiner1989}.

\subsubsection{Models for Both Hands}
\label{sec:ModelForBothHands}

A piano-score model with the left and right hand parts can be obtained by first constructing a model for each hand part and then combining the two models.
If musical notes are already assigned to two hand parts, such a combined model can be obtained directly.
On the other hand, if the part assignment is not given, as in the piano reduction problem, the model should be able to describe the probability for all cases of part assignment.

Such a model for piano music with unknown hand parts can be constructed based on the merged-output HMM \cite{Nakamura2014,Nakamura2017TASLP}.
The idea is to combine outputs from two component Markov models or HMMs respectively describing the two hand parts.
We here describe a model combining two fingering models.
First, the hand part (left or right) associated with a note $p_n$ is represented by an additional stochastic variable $\eta_n\in\{{\rm L},{\rm R}\}$.
The generative process of $\eta_n$ is described with a Bernoulli distribution: $P(\eta_n=\eta)=\alpha_{\eta}$ with $\alpha_{\rm L}+\alpha_{\rm R}=1$.
If $\eta_n$ is determined, then the pitch is generated by the corresponding component model.
For each $\eta\in\{{\rm L},{\rm R}\}$, let $a^\eta_{f'f}=P^\eta(f|f')$ and $b^\eta_{f'f}(p',p)=F^\eta(\bm\ell(p)-\bm\ell(p');f',f)$ denote the fingering and pitch transition probabilities of the component model.
This process can be described as an HMM with a state space indexed by $k=(\eta,f^{\rm L},p^{\rm L},f^{\rm R},p^{\rm R})$ with the following initial and transition probabilities:
\begin{align}
&P(k_n=k\,|\,k_{n-1}=k')
\notag\\
&=
\begin{cases}
\alpha_{\rm L}a^{\rm L}_{f'^{\rm L}f^{\rm L}}b^{\rm L}_{f'^{\rm L}f^{\rm L}}(p'^{\rm L},p^{\rm L})\delta_{f'^{\rm R}f^{\rm R}}\delta_{p'^{\rm R}p^{\rm R}}, &\eta={\rm L};\\
\alpha_{\rm R}a^{\rm R}_{f'^{\rm R}f^{\rm R}}b^{\rm R}_{f'^{\rm R}f^{\rm R}}(p'^{\rm R},p^{\rm R})\delta_{f'^{\rm L}f^{\rm L}}\delta_{p'^{\rm L}p^{\rm L}}, &\eta={\rm R},\\
\end{cases}
\notag\\
&P(p_n=p\,|\,k_n=k)=\delta_{pp^\eta},
\end{align}
where $\delta$ denotes Kronecker's delta.

Using this model, one can estimate the sequence of latent variables $k_{1:N}$ from a pitch sequence $p_{1:N}$.
This can be done by maximizing the probability $P(k_{1:N}\,|\,p_{1:N})\propto P(k_{1:N},p_{1:N})$ 
The most probable sequence $\hat{k}_{1:N}$ has the information of the optimal configuration of hands $\hat{\eta}_{1:N}$, which yields separated two hand parts and the optimal fingering for both hands ($\hat{f}^{\rm L}_{1:N}$ and $\hat{f}^{\rm R}_{1:N}$).
For more details, see \cite{Nakamura2014}.
The Gaussian model and the no-information model can be similarly extended to models for both hands.

\subsection{Performance Difficulty}
\label{sec:Difficulty}

\subsubsection{Difficulty Measures}

One can define a quantitative measure of performance difficulty based on the cost of music performance.
From the statistical viewpoint, a natural choice is the probabilistic cost, which is the negative logarithm of a probability.
To include the dependence on tempo, we define a performance difficulty as the time rate of the probabilistic cost
\begin{equation}
D(t)=-{\rm ln}\,P(\bm p(t))/\Delta t.
\end{equation}
Here, $\Delta t$ is a time width, $\bm p(t)$ is the sequence of pitches in the time range $[t-\Delta t/2,t+\Delta t/2]$, and $P(\bm p(t))$ is defined with one of the piano-score models in section \ref{sec:FingeringModelAndDifficulty}.\ref{sec:FingeringModel}.
With the fingering model, one can use the joint probability of pitches and fingering to define a difficulty measure \cite{Nakamura2014}:
\begin{equation}
D(t)=-{\rm ln}\,P(\bm p(t),\bm f(t))/\Delta t,
\label{eq:DiffWithFingering}
\end{equation}
where $\bm f(t)$ denotes the fingering corresponding to the pitches $\bm p(t)$.
If the fingering is unknown, one can substitute the maximum-probability estimate $\hat{\bm f}(t)$ in Eq.~(\ref{eq:DiffWithFingering}).
For each note $n$ with onset time $t_n$, we write $D(n)=D(t_n)$.

The difficulty measure can be defined for each hand part using the pitches in that hand part and a piano-score model for one hand, which is denoted by $D_{\rm L}(t)$ or $D_{\rm R}(t)$.
In addition, the total difficulty can be defined as the sum of difficulties for both hands: $D_{\rm B}(t)=D_{\rm L}(t)+D_{\rm R}(t)$.
The quantity $D_{\rm B}(t)$ can be relevant as well as $D_{\rm L}(t)$ and $D_{\rm R}(t)$ since the difficulty can be high even if difficulties for individual hand parts are not so high.

In previous studies \cite{Chiu2012,Sebastien2012}, features such as playing speed, note density, pitch entropy, hand displacement rate, hand stretch, and fingering complexity have been considered to estimate the difficulty level of piano scores.
These features are incorporated in the above difficulty measures, although in an implicit manner.
If one uses the no-information model, the difficulty measure takes into account the note density and playing speed.
With the Gaussian model, pitch entropy and hand displacement rate, and hand stretch are incorporated in addition.
With the fingering model, fingering complexity is further incorporated.

\subsubsection{Evaluation}
\label{sec:DifficultyEvaluation}

To formally examine how the proposed measures reflect real performance difficulty, we study their relation with the rate of performance errors.
We use a dataset \cite{Nakamura2017} consisting of 90 MIDI piano performance signals of 30 classical musical pieces; for each piece there are performances by three different players that are recorded in international piano competitions.
In the dataset, musical notes in a performance signal are matched to notes in the corresponding score and the following three types of performance errors are manually annotated: pitch error (a performed note with a corresponding note in the score but with a different pitch); extra note (a performed note without a corresponding note in the score); and missing note (a note in the score without a corresponding note in the performance).
Timing errors are not annotated in the data and not considered in this study.

We calculate performance difficulty values for each onset time and calculate the number of performance errors in the time range of width $\Delta t$ around the onset time.
In the following, we set $\Delta t$ to be $1$ s.
For the Gaussian model, $\epsilon=4\times10^{-4}$ and $p_0$ is C3 (C5) for the left (right) hand.
Other parameters of the Gaussian and fingering models are taken from a previous study \cite{Nakamura2014} where a different dataset was used for training.

\begin{figure}[t]
\centering
{\includegraphics[clip,width=0.95\columnwidth]{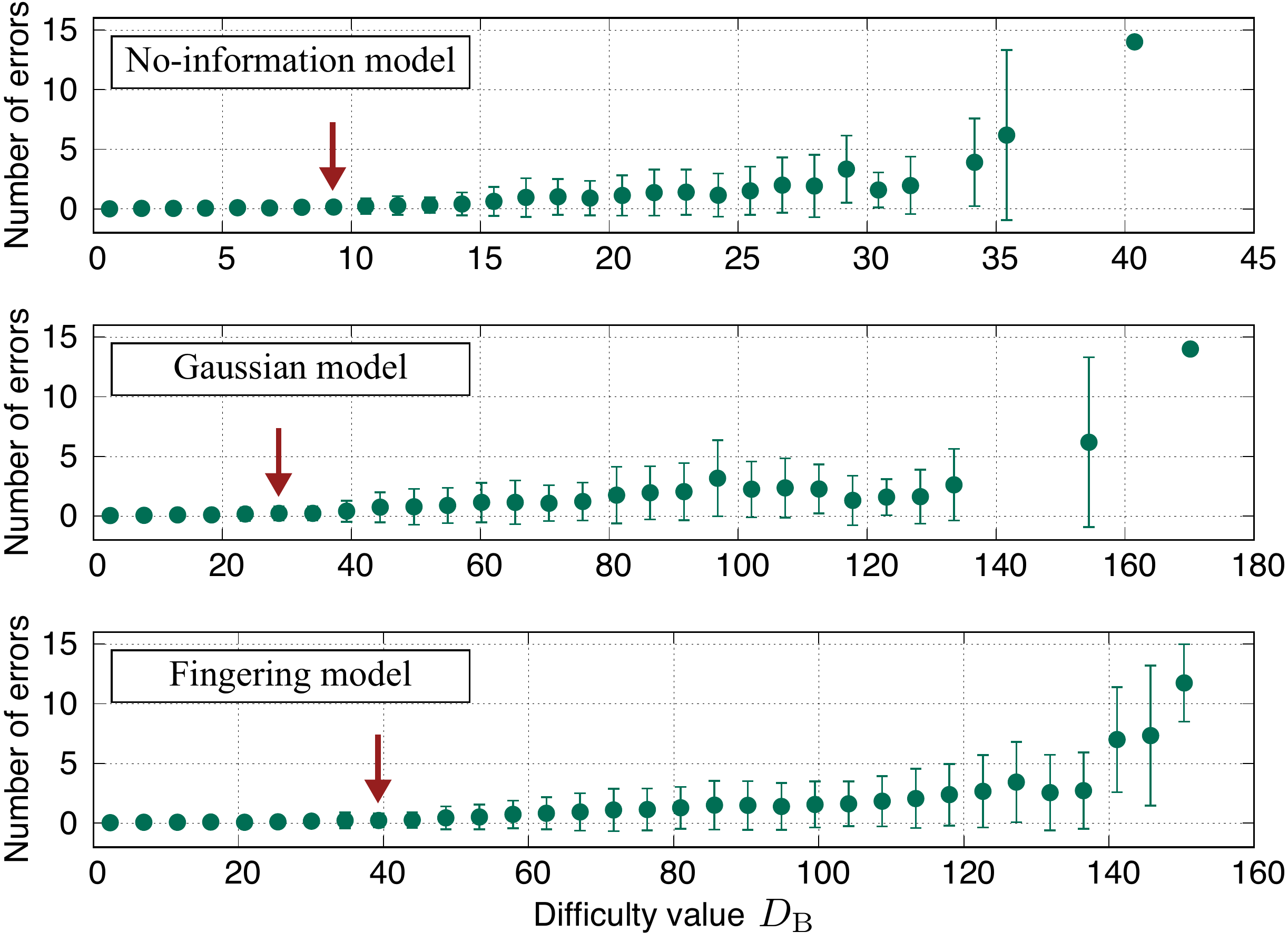}}
\caption{Relations between the difficulty value $D_{\rm B}$ and the number of performance errors. Points and bars indicate means and standard deviations. Arrows indicate onsets of performance errors (see text).}
\label{fig:DiffErr}
\end{figure}
Fig.~\ref{fig:DiffErr} shows the relation between difficulty value $D_{\rm B}$ and the rate of performance errors for the three models.
We see that for each model there is an onset (roughly, $10$ for the no-information model, $30$ for the Gaussian model, and $40$ for the fingering model) below which the average number of errors is almost zero and above which it gradually increases.
This suggests that the difficulty measures can be used as indicators of performance errors.

For comparative evaluation, we predict performance errors by thresholding the difficulty values and measure the predictive accuracy for the three models.
Using three thresholds $D^{\rm th}_{\rm L}$, $D^{\rm th}_{\rm R}$, and $D^{\rm th}_{\rm B}$, a prediction of performance errors at time $t$ is defined positive if one of three conditions ($D^{\rm th}_{\rm L}>D_{\rm L}(t)$, $D^{\rm th}_{\rm R}>D_{\rm R}(t)$, and $D^{\rm th}_{\rm B}>D_{\rm B}(t)$) is satisfied.
We calculate the number of true positives $N_{\rm TP}$, that of false positives $N_{\rm FP}$, and that of true negatives $N_{\rm TN}$, and the following quantities are used as evaluation measures:
\begin{equation}
\mathcal{P}=\frac{N_{\rm TP}}{N_{\rm TP}+N_{\rm FP}},\quad \mathcal{R}=\frac{N_{\rm TP}}{N_{\rm TP}+N_{\rm TN}},\quad \mathcal{F}=\frac{2\mathcal{P}\mathcal{R}}{\mathcal{P}+\mathcal{R}}.
\notag
\end{equation}
Since more frequent errors indicate larger difficulty, we can also define the following weighted quantities:
\begin{align}
\mathcal{P}_{\rm w}&=\frac{N'_{\rm TP}}{N'_{\rm TP}+N_{\rm FP}},\quad \mathcal{R}_{\rm w}=\frac{N'_{\rm TP}}{N'_{\rm TP}+N'_{\rm TN}},
\\
\mathcal{F}_{\rm w}&=\frac{2\mathcal{P}_{\rm w}\mathcal{R}_{\rm w}}{\mathcal{P}_{\rm w}+\mathcal{R}_{\rm w}},
\end{align}
where $N'_{\rm TP}$ and $N'_{\rm TN}$ are obtained by weighting $N_{\rm TP}$ and $N_{\rm TN}$ with the number of performance errors.
The results are shown in Table \ref{tab:DifficultyEvaluation} where the thresholds are optimized with respect to $\mathcal{F}_{\rm w}$ for each model.
We see that the Gaussian model has the highest F measures, even though the differences are rather small.
A possible reason is the relatively small size of the data used for training the fingering model.
Since the Gaussian model has only one parameter $\sigma_p$ to train, it has better generalization ability for such small training data.
Such a trade-off between model complexity and the required amount of training data is common in many machine-learning problems.
We thus use the difficulty measures defined with the Gaussian model in the following.
\begin{table*}[t]
\footnotesize
\centering
\begin{tabular}{ccllllll}\toprule
Model & Threshold $(D^{\rm th}_{\rm R},D^{\rm th}_{\rm L},D^{\rm th}_{\rm B})$ & $\mathcal{F}$ & $\mathcal{P}$ & $\mathcal{R}$ & $\mathcal{F}_{\rm w}$ & $\mathcal{P}_{\rm w}$ & $\mathcal{R}_{\rm w}$\\
\midrule
No-information  & $(9,10,14)$ & $52.4$ & $43.0$ & $67.1$ & $69.8$ & $63.0$ & $78.1$\\
Gaussian        & $(30,30,42)$ & $54.2$ & $46.3$ & $65.2$ & $71.3$ & $66.4$ & $77.0$\\
Fingering       & $(41,39,53)$ & $53.9$ & $49.1$ & $59.8$ & $70.6$ & $69.3$ & $73.8$\\
\bottomrule
\end{tabular}
\vspace{2mm}
\caption{Accuracies of performance error prediction.}
\label{tab:DifficultyEvaluation}
\vspace{-2mm}
\end{table*}
%

\section{Piano Reduction Method}
\label{sec:Method}

In the statistical formulation of piano reduction, we try to find the optimal reduction score $\hat{R}$ that maximizes the probability $P(R|E)$ for a given ensemble score $E$.
In analogy with the statistical approach for machine translation \cite{StatisticalTranslation}, we first construct generative models describing the probability $P(R)$ and $P(E|R)$ respectively and integrate them for calculating $P(R,E)\propto P(R|E)$.
We then derive optimization algorithms for piano reduction that take into account the constraints on performance difficulty values.

Prior to the main processing step, we convert an input ensemble score to a condensed score by removing redundant notes with the same pitch and simultaneous onset time (the number of such redundant notes is memorized and used later in the calculation of Eq.~(\ref{eq:Importance})).
Although they are different strictly, we call such a condensed score an ensemble score in what follows.
What is really meant by the symbol $E$ is also a condensed score.

\subsection{Model for Piano Reduction}
\label{sec:ModelForPianoReduction}

To construct a generative model that yields the probability $P(R,E)$, we integrate a piano-score model describing the probability $P(R)$ and an {\it edit model} that describes the process yielding $P(E|R)$.
As a piano-score model, we can use either the Gaussian model or the fingering model discussed in section \ref{sec:FingeringModelAndDifficulty}.\ref{sec:FingeringModel}.\ref{sec:ModelForBothHands}, which statistically describes the naturalness of a generated (reduction) score.

For the edit model, we assign probabilities for edit operations applied to musical notes.
As in \cite{Nakamura2015}, we focus on the two most common edit operations, note deletion and octave pitch shift.
As we model the inverse process of generating an ensemble score from a piano score, we introduce probabilities of note addition and octave pitch shift in the edit model.
For each note in the ensemble score, the probability that it is an added note and not originated from the piano score is denoted by $\beta_{\rm NP}$ (`NP' for not played).
In this case, the note's pitch $p$ is drawn from a uniform distribution $c_{\rm unif}(p)$.
If it is originated from the piano score and the corresponding note has a pitch $q$, the probability of the note's pitch $p$ denoted by $c_q(p)=P(p|q)$ is supposed to obey
\begin{equation}
c_q(p)=
\begin{cases}
1-2\gamma_{\rm oct}, & p=q;\\
\gamma_{\rm oct}, &p=q\pm12,
\end{cases}
\end{equation}
where $\gamma_{\rm oct}$ denotes the probability of an octave shift.

\begin{figure}[t]
\centering
{\includegraphics[clip,width=0.97\columnwidth]{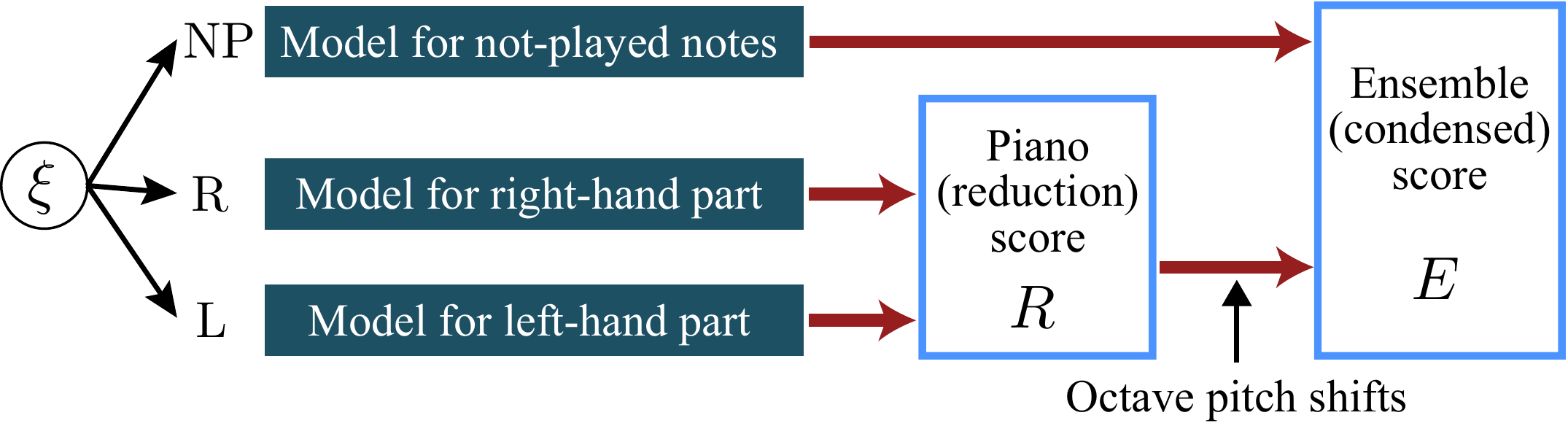}}
\vspace{-2mm}
\caption{Generative process of the model for piano reduction.}
\label{fig:PianoReductionModel}
\end{figure}
We can integrate the fingering model in section \ref{sec:FingeringModelAndDifficulty}.\ref{sec:FingeringModel}.\ref{sec:ModelForBothHands} and the edit model in the following fashion based on the merged-output HMM (Fig.~\ref{fig:PianoReductionModel}), which leads to tractable inference algorithms.
For each note in the output ensemble score, indexed by $m$, we introduce a stochastic variable $\xi_m$ that can take one of three values $\{{\rm NP},{\rm L},{\rm R}\}$.
It is generated from a discrete distribution as $P(\xi_m=\xi)=\beta_{\xi}$, where parameters $\beta_{\xi}$ obey $\beta_{\rm NP}+\beta_{\rm L}+\beta_{\rm R}=1$.
If $\xi_m={\rm NP}$, then its pitch $p_m$ has a probability $P(p_m=p)=c_{\rm unif}(p)$.
If $\xi_m={\rm L}$ or ${\rm R}$, then its pitch is generated from the component fingering model of the corresponding hand part and may undergo an octave shift.
Writing $f^{\rm L}$, $p^{\rm L}$, $f^{\rm R}$, and $p^{\rm R}$ for the finger and pitch variables of the two component fingering models, the latent state of the merged-output HMM is described by a set of variables $r=(\xi,f^{\rm L},p^{\rm L},f^{\rm R},p^{\rm R})$.
The transition and output probabilities are defined as
\begin{align}
&P(r_m=r\,|\,r_{m-1}=r')
\notag\\
&~~=
\begin{cases}
\beta_{\rm NP}\delta_{f^{\rm L}f'^{\rm L}}\delta_{f^{\rm R}f'^{\rm R}}\delta_{p^{\rm L}p'^{\rm L}}\delta_{p^{\rm R}p'^{\rm R}}, &\xi={\rm NP};\\
\beta_{\rm L} a^{\rm L}_{f'^{\rm L}f^{\rm L}}b^{\rm L}_{f'^{\rm L}f^{\rm L}}(p'^{\rm L},p^{\rm L})\delta_{f^{\rm R}f'^{\rm R}}\delta_{p^{\rm R}p'^{\rm R}}, &\xi={\rm L};\\
\beta_{\rm R} a^{\rm R}_{f'^{\rm R}f^{\rm R}}b^{\rm R}_{f'^{\rm R}f^{\rm R}}(p'^{\rm R},p^{\rm R})\delta_{f^{\rm L}f'^{\rm L}}\delta_{p^{\rm L}p'^{\rm L}}, &\xi={\rm R},
\end{cases}
\notag
\\
&P(p_m=p\,|\,r_m=r)=
\begin{cases}
c_{\rm unif}(p), &\xi={\rm NP};\\
c_{p^{\rm L}}(p), &\xi={\rm L};\\
c_{p^{\rm R}}(p), &\xi={\rm R}.
\end{cases}
\end{align}
The model indeed generates a piano score specified by $(p^{\rm L}_m,p^{\rm R}_m)_m$ and an ensemble score specified by $(p_m)_m$.

In the process of piano reduction, which is explained in the next section, the parameter $\beta_{\rm NP}$ represents how much notes in the ensemble score are removed.
Thus, properly adjusting $\beta_{\rm NP}$ is crucial to control the performance difficulty of resulting reduction scores.
Roughly speaking, if the note density around a note is high, it is necessary to remove more notes around that note by setting $\beta_{\rm NP}$ large.
In addition, some notes like melodic notes and bass notes are musically more important than others and should have a small probability of deletion, or small $\beta_{\rm NP}$ in the present model.
These conditions can be realized in the following form of $\beta_{\rm NP}(m)$, which depends on each note $m$:
\begin{equation}
\beta_{\rm NP}(m)=\big(1-\zeta(m)\big)e^{-\kappa h(m)},
\label{eq:BetaNP}
\end{equation}
where $h(m)\geq0$ represents the musical importance of note $n$, $\kappa>0$ is a coefficient to control the effect of $h(m)$, and $\zeta(m)\in[0,1]$ is a factor to control the overall rate of note deletion.
If $\zeta(m)\simeq1$, $\beta_{\rm NP}(m)\simeq0$ and almost all notes remain in the reduction score.
If $\zeta(m)\simeq0$, $\beta_{\rm NP}(m)\simeq1$ unless $\kappa h(m)$ is large (i.e.\ note $m$ is musically important), so most musically unimportant notes will be removed.

In addition to the importance of melodic and bass notes, it is not difficult to imagine that pitches in an ensemble score that are played simultaneously by multiple instruments are musically important.
Thus, the following form is used for defining musical importance $h(m)$:
\begin{equation}
h(m)=\mathbb{I}(m\in\mathcal{M})+\mathbb{I}(m\in\mathcal{B})+a\,{\rm Mult}(m),
\label{eq:Importance}
\end{equation}
where $\mathbb{I}(\mathcal{C})=1$ if a condition $\mathcal{C}$ is true and $0$ otherwise, $\mathcal{M}$ denotes the set of melodic notes, $\mathcal{B}$ denotes the set of bass notes, and ${\rm Mult}(m)$ is the {\it multiplicity} of note $m$, defined as the number of notes in the ensemble score having the same pitch and onset time as note $m$ excluding $m$ itself.
The parameters $\kappa$ and $a$ are adjustable parameters, and $\zeta(m)$ is adjusted according to target difficulty values as explained in the next section.

\subsection{Algorithms for Piano Reduction}
\label{sec:ReductionAlgorithm}

Let us derive algorithms for piano reduction based on the model in section \ref{sec:Method}.\ref{sec:ModelForPianoReduction} and the difficulty measures in section \ref{sec:FingeringModelAndDifficulty}.\ref{sec:Difficulty}.
The piano reduction problem is here formulated as finding a reduction score $R$ that maximizes $P(R|E)$ for a given ensemble score $E$ with constraints on $R$'s performance difficulty values.
Specifically, we impose the following constraints for each note $n$ in $R$:
\begin{equation}
[D_{\rm L}(n)<\widetilde{D}_{\rm L}]\,\wedge\,[D_{\rm R}(n)<\widetilde{D}_{\rm R}]\,\wedge\,[D_{\rm B}(n)<\widetilde{D}_{\rm B}],
\label{eq:DifficultyConstraints}
\end{equation}
where $\widetilde{D}_{\rm L}$, $\widetilde{D}_{\rm R}$, and $\widetilde{D}_{\rm B}$ are some target difficulty values.

Without the constraints (\ref{eq:DifficultyConstraints}), finding the maximum of $P(R|E)$ is a basic inference problem for HMMs and can be achieved with the Viterbi algorithm \cite{Rabiner1989}.
However, the constraints (\ref{eq:DifficultyConstraints}) cannot be easily treated because difficulty values at each note depends on the existence of other notes in the time range of $\Delta t$, which violates the Markovian assumption for the Viterbi algorithm.
In other words, if we know appropriate values of $\zeta(m)$ in Eq.~(\ref{eq:BetaNP}) for controlling difficulty values, the optimization problem is directly solvable, but finding those values is not easy.

In the following, we present two strategies for optimization.
In a previous study \cite{Nakamura2015}, appropriate values of $\zeta(m)$ were estimated and the Viterbi algorithm was applied once to obtain the result.
A slight extension of this {\it one-time optimization} method is presented in section \ref{sec:Method}.\ref{sec:ReductionAlgorithm}.\ref{sec:OneTimeOptimization}.
On the other hand, if one can apply the Viterbi algorithm iteratively, it would be possible to find appropriate values of $\zeta(m)$ from tentative results, by starting from $\zeta(m)=1$ and gradually lessening it.
This {\it iterative optimization} method is developed in section \ref{sec:Method}.\ref{sec:ReductionAlgorithm}.\ref{sec:IterativeOptimization}.

\subsubsection{One-time optimization algorithm}
\label{sec:OneTimeOptimization}

In \cite{Nakamura2015}, appropriate values of $\zeta(m)$ were estimated by matching the expected difficulty values to the target values with the following equation:
\begin{equation}
\zeta(m)={\rm min}\bigg\{\frac{\widetilde{D}_{\rm L}}{D_{\rm L}(m)},\frac{\widetilde{D}_{\rm R}}{D_{\rm R}(m)}\bigg\},
\label{eq:ExpZetaBasic}
\end{equation}
where $D_{\rm L}(m)$ etc.\ represent the difficulty values calculated for the ensemble score at its $m$\,th note.
One can include a factor involving $D_{\rm B}(m)$ in the above equation in general.

We can generalize this method by introducing a scaling factor $\rho$ and modifying Eq.~(\ref{eq:ExpZetaBasic}) to
\begin{equation}
\zeta(m)=\rho\,{\rm min}\bigg\{\frac{\widetilde{D}_{\rm L}}{D_{\rm L}(m)},\frac{\widetilde{D}_{\rm R}}{D_{\rm R}(m)}\bigg\}.
\label{eq:ExpZetaGeneral}
\end{equation}
By choosing the value of $\rho$, one can control the expected average of resulting difficulty values.
For example, one can use a maximum value of $\rho$ that can satisfy the constraints (\ref{eq:DifficultyConstraints}) for most outcomes.

\subsubsection{Iterative optimization algorithm}
\label{sec:IterativeOptimization}

For iterative optimization, the Viterbi algorithm is applied in each iteration to obtain a tentative reduction score $R^{(i)}$, with tentative values of $\zeta^{(i)}(m)$ ($i$ is an index for iterations).
For each note $n$ in $R^{(i)}$, we calculate the difficulty values $D^{(i)}_{\rm L}(n)$, $D^{(i)}_{\rm R}(n)$, and $D^{(i)}_{\rm B}(n)$.
If the constraints (\ref{eq:DifficultyConstraints}) are not all satisfied at note $n$, then we lessen the values of $\zeta(m)$ for all notes $m$ in the ensemble score around $n$ within the time range of width $\Delta t$ as
\begin{equation}
\zeta^{(i+1)}(m)=\lambda\zeta^{(i)}(m)
\label{eq:UpdateControlFactor}
\end{equation}
with some constant $0<\lambda<1$.

The iterative algorithm is initialized with $\zeta^{(i=1)}(m)=1$ for all notes $m$.
The algorithm ends when the constraints (\ref{eq:DifficultyConstraints}) are satisfied at every note in the reduction score, or the number of iterations exceed some predefined value $i_{\rm max}$.
For efficient and stable computation, the Viterbi algorithm at iteration $i+1$ is applied only to those regions of the ensemble score where the constraints (\ref{eq:DifficultyConstraints}) are not still satisfied at iteration $i$.
Specifically, we first construct a set of notes $m$ in the ensemble score whose onset time $t_m$ is included in the range $[t_n-\Delta t/2,t_n+\Delta t/2]$ around some onset time $t_n$ in the reduction score for which the difficulty constraints are not satisfied.
This set is then split into a set $\Psi$ of isolated regions of notes.
For each such isolated region, the Viterbi algorithm is applied with fixed boundary states at one note before the beginning of the region and one note after the end.

The iterative algorithm is summarized as follows.
\begin{enumerate}
\item Initialize $\zeta^{(i=1)}(m)=1$ and apply the Viterbi algorithm to the whole ensemble score.
\item Calculate difficulty values and obtain regions $\Psi$ where the constraints (\ref{eq:DifficultyConstraints}) are not satisfied. Exit if $\Psi$ is empty or $i\geq i_{\rm max}$.
\item Update the control factor $\zeta(m)$ as in Eq.~(\ref{eq:UpdateControlFactor}) and apply the Viterbi algorithm to each region of $\Psi$. Increment $i$ and go back to step (ii).
\end{enumerate}
%

\section{Evaluation of Piano Reduction Algorithms}
\label{sec:Evaluation}

\subsection{Setup}

To evaluate the piano reduction algorithms, we prepared a dataset of orchestral pieces of Western classical music.
The dataset consists of $10$ pieces by different composers and with different instrumentations; each pieces has a length of around 20 bars.
The list of the pieces are available in the accompanying webpage \footnote{\url{http://pianoarrangement.github.io/demo.html}}.

We compare one-time optimization algorithms and iterative optimization algorithms based on the Gaussian model and the fingering model; in total we have four methods labelled as {\it One-time Gaussian}, {\it One-time Fingering}, {\it Iterated Gaussian}, and {\it Iterated Fingering} methods.
The parameters of the piano-score models are set as in section \ref{sec:FingeringModelAndDifficulty}.\ref{sec:Difficulty}.\ref{sec:DifficultyEvaluation}.
The other parameters are set as follows: $a=0.01$, $\kappa=11$, $\lambda=0.85$, $\gamma_{\rm oct}=0.001$, and $\beta_{\rm R}(m)=\beta_{\rm L}(m)=(1-\beta_{\rm NP}(m))/2$ where $\beta_{\rm NP}(m)$ is set as in Eq.~(\ref{eq:BetaNP}).
Difficulty values are calculated with the difficulty measures using the Gaussian model with $\Delta t=1$ s.
These parameter values were fixed after some trials by one of the authors and there is room for further optimization.

As a baseline method we also implement a method based on a simple piano-score model (called the {\it distance} model) that takes into account the distance between each note in the ensemble score and its closest melodic or bass note, but not sequential dependence of pitches.
Specifically, for each note $m$ in the ensemble score the closest melodic or bass notes ${\rm CMB}(m)$ is obtained by first searching in the direction of onset time and then in the direction of pitches.
Then the probability of its pitch $p_m$ is given as
\begin{equation}
P(p_m)\propto {\rm Gauss}(p_m;{\rm CMB}(m),\sigma_p^2).
\label{eq:RandomPitchProb}
\end{equation}
Integrating this piano-score model into the piano reduction model in section \ref{sec:Method}.\ref{sec:ModelForPianoReduction} and using the iterative optimization algorithm, a baseline {\it Iterated Distance} method is obtained.

\subsection{Quantitative Evaluation of Difficulty Control}

We first examine the effect of the iterative optimization algorithms in controlling the difficulty values of output reduction scores, in comparison with the one-time optimization algorithms.
We run the four algorithms, {\it One-time Gaussian}, {\it One-time Fingering}, {\it Iterated Gaussian}, and {\it Iterated Fingering}, for the test dataset with three sets of target difficulty values $(\widetilde{D}_{\rm L},\widetilde{D}_{\rm R},\widetilde{D}_{\rm B})=(15,15,30)$, $(30,30,40)$, and $(40,40,50)$.
For the scaling factor $\rho$ for the one-time optimization algorithms, we test values in $\{0.1,0.2,\ldots,1.0\}$.
For the iterative optimization algorithms, $i_{\rm max}$ is set to $50$.
To evaluate a reduction score $R$, we compute difficulty values ($D_{\rm L}(n)$ etc.) for each note $n$ in $R$ and calculate the following measures:
\begin{itemize}
\item Mean Difficulty Values $\{\overline{D}_{\rm L},\overline{D}_{\rm R},\overline{D}_{\rm B}\}$:
\begin{equation}
\overline{D}_{\rm L}=\frac{1}{\#R}\sum_{n\in R}D_{\rm L}(n)\quad{\rm etc.}
\end{equation}
\item Maximum Difficulty Values $\{D^{\rm max}_{\rm L},D^{\rm max}_{\rm R},D^{\rm max}_{\rm B}\}$:
\begin{equation}
D^{\rm max}_{\rm L}=\max_{n\in R}\{D_{\rm L}(n)\}\quad{\rm etc.}
\end{equation}
\item Out-of-Range Rate (proportion of regions where difficulty values exceed target values) $\{A_{\rm out}^{\rm L},A_{\rm out}^{\rm R},A_{\rm out}^{\rm B}\}$:
\begin{equation}
A_{\rm out}^{\rm L}=\frac{\#\big\{n\in R\big|D_{\rm L}(n)>\widetilde{D}_{\rm L}\big\}}{\#R}\quad{\rm etc.}
\end{equation}
\item Additional-Note Rate (proportion of notes in the reduction score other than melodic and bass notes) $A_{\rm add}$:
\begin{equation}
A_{\rm add}=\frac{\#R-\#\mathcal{M}-\#\mathcal{B}}{\#\mathcal{M}+\#\mathcal{B}}.
\end{equation}
\end{itemize}
\begin{figure}[t]
\centering
{\includegraphics[clip,width=0.95\columnwidth]{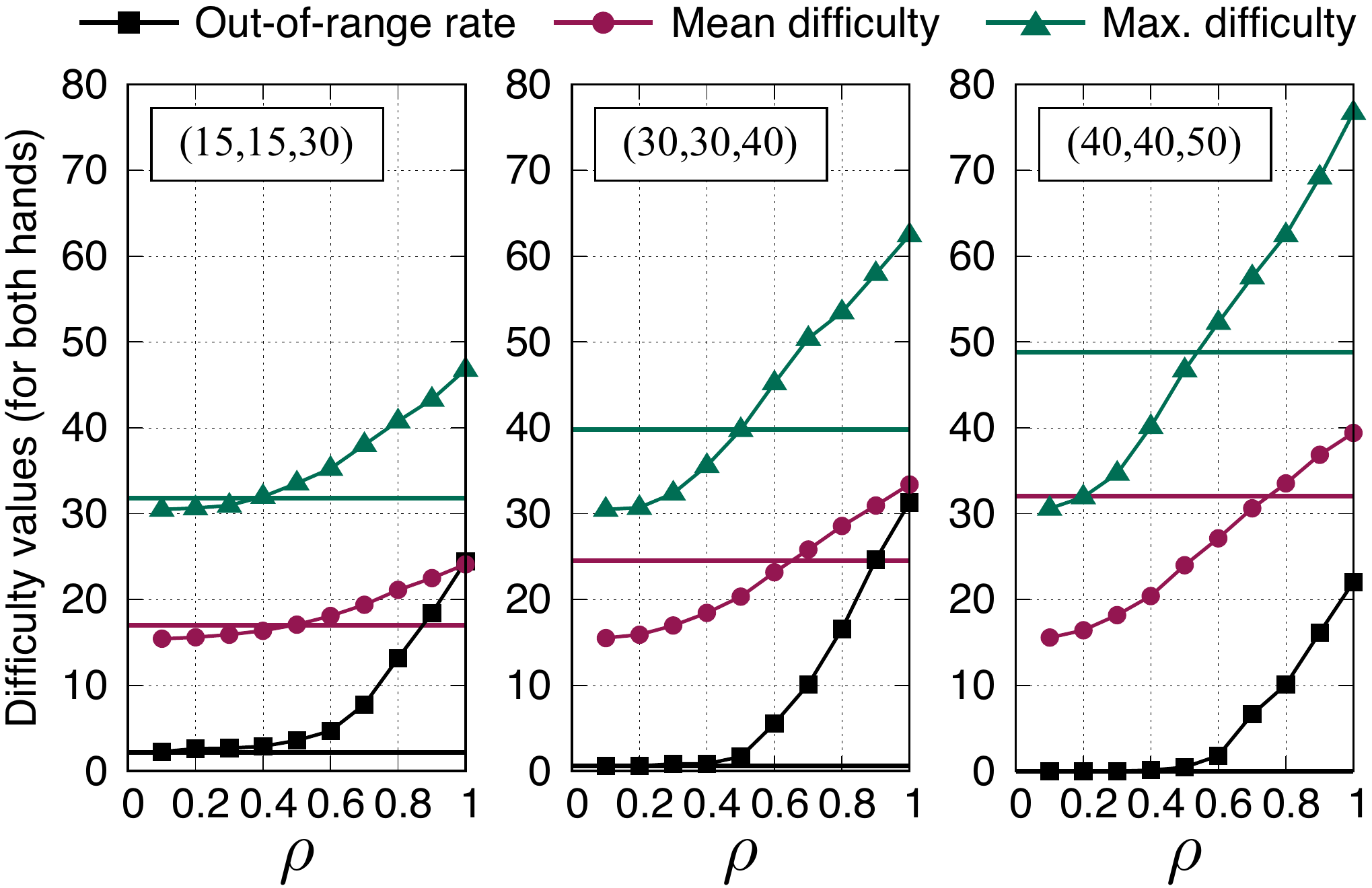}}
\caption{Difficulty metrics for the {\it One-time Gaussian} method for varying $\rho$, for three cases of target difficulty values $(\widetilde{D}_{\rm L},\widetilde{D}_{\rm R},\widetilde{D}_{\rm B})$ indicated in the insets. Difficulty values are those for both hands ($\overline{D}_{\rm B}$, $D^{\rm max}_{\rm B}$, etc.) and horizontal lines indicate corresponding values for the {\it Iterated Gaussian} method.}
\label{fig:DifficultyControl}
\end{figure}
Variations of difficulty values of the reduction scores by the {\it One-time Gaussian} method are shown in Fig.~\ref{fig:DifficultyControl}, with corresponding values for the {\it Iterated Gaussian} method.
Here, for simplicity, only difficulty values for both hands ($\overline{D}_{\rm B}$, $D^{\rm max}_{\rm B}$, etc.) are shown.
It is observed that for those values of $\rho$ where $A^{\rm B}_{\rm out}$ is equivalent to that for the iterative optimization method, $\overline{D}_{\rm B}$ and $D^{\rm max}_{\rm B}$ are smaller compared to the iterative optimization method.
This means that with the same level of satisfaction for the difficulty constraints, results of the iterative optimization method have larger difficulty values on the average, which is a desired property.
On the other hand, if $\rho$ is increased sufficiently, it is possible for the one-time optimization algorithm to achieve the same level of $\overline{D}_{\rm B}$ as the iterative optimization method, but then $A^{\rm B}_{\rm out}$ is larger, meaning that the difficulty constraints are less strictly satisfied.
Analyses of difficulty values for each hand and comparison between {\it One-time Fingering} and {\it Iterated Fingering} methods reveal similar tendencies.

\begin{table*}[t]
\footnotesize
\centering
\begin{tabular}{crrrrr}\toprule
Algorithm          & Target difficulty & Mean difficulty   & Max.\ difficulty   & Out-of-range rate (\%) & $A_{\rm out}$ (\%) \\
\midrule
One-time Gaussian  & $(15,15,30)$      & $(10.0,5.4,15.4)$ & $(22.5,14.6,30.5)$ & $(18.6,7.3,2.3)$   & $7.1$ \\
Iterated Gaussian  & $(15,15,30)$      & $(11.0,6.1,17.0)$ & $(22.3,15.5,31.8)$ & $(18.2,7.2,2.2)$   & $20.3$ \\
One-time Fingering & $(15,15,30)$      & $(12.9,9.1,22.0)$ & $(30.7,27.8,50.5)$ & $(30.0,18.0,20.9)$ & $30.9$ \\
Iterated Fingering & $(15,15,30)$      & $(12.7,8.4,21.1)$ & $(29.0,23.9,46.5)$ & $(27.5,15.7,14.9)$ & $31.7$ \\
Iterated Distance    & $(15,15,30)$      & $(11.9,6.1,18.0)$ & $(28.0,15.7,37.3)$ & $(23.4,7.4,5.2)$   & $21.8$ \\
\midrule
One-time Gaussian  & $(30,30,40)$      & $(10.4,5.5,15.9)$ & $(23.2,15.4,30.7)$ & $(0.7,0,0.6)$      & $11.4$ \\
Iterated Gaussian  & $(30,30,40)$      & $(16.2,8.3,24.5)$ & $(30.0,21.2,39.8)$ & $(0.4,0,0.6)$      & $62.3$ \\
One-time Fingering & $(30,30,40)$      & $(13.2,9.4,22.7)$ & $(31.8,28.6,51.2)$ & $(6.5,5.8,11.6)$   & $33.4$ \\
Iterated Fingering & $(30,30,40)$      & $(16.3,10.6,26.9)$& $(34.3,28.6,50.9)$ & $(3.6,3.0,6.3)$    & $60.1$ \\
Iterated Distance    & $(30,30,40)$      & $(17.8,8.3,26.0)$ & $(35.9,21.7,44.8)$ & $(2.4,0,2.3)$      & $61.9$ \\
\midrule
One-time Gaussian  & $(40,40,50)$      & $(13.4,7.0,20.4)$ & $(30.6,19.2,40.1)$ & $(0.1,0,0.1)$      & $39.0$ \\
Iterated Gaussian  & $(40,40,50)$      & $(20.9,11.1,32.0)$& $(36.8,27.8,48.8)$ & $(0,0,0)$          & $98.3$ \\
One-time Fingering & $(40,40,50)$      & $(13.5,9.5,22.9)$ & $(32.4,29.2,51.7)$ & $(2.8,3.4,5.7)$    & $34.7$ \\
Iterated Fingering & $(40,40,50)$      & $(20.2,13.6,33.8)$& $(40.1,33.1,54.9)$ & $(1.7,1.0,1.6)$    & $88.9$ \\
Iterated Distance    & $(40,40,50)$      & $(22.1,10.3,32.4)$& $(42.5,27.3,53.6)$ & $(0.8,0,0.8)$      & $88.3$ \\
\bottomrule
\end{tabular}
\vspace{2mm}
\caption{Comparison of average values of difficulty metrics for reduction scores. Triplet values in parentheses indicate one for left-hand part, right-hand part, and both hand parts, from left to right.}
\label{tab:ObjectiveEvaluation}
\vspace{-2mm}
\end{table*}
The results for all three kinds of difficulty values (for each of two hands and for both hands) are shown in Table \ref{tab:ObjectiveEvaluation}.
Here, for one-time optimization methods, results are shown for the smallest value of $\rho$ such that all three out-of-range rates exceed those for the corresponding iterative optimization methods.
In addition to the same tendencies as found in the above analysis, one can observe that for the same level of satisfaction of difficulty constraints, the iterative optimization methods yields larger additional-note rates than the corresponding one-time optimization methods.
These results indicate that the iterative optimization methods are more appropriate for controlling difficulty values.

Even for iterative optimization algorithms, the out-of-range rates can be nonzero especially for small target difficulty values.
One reason for this is that for some pieces the minimal reduction score with only melodic and bass notes has difficulty values larger than the target values.
Another reason is the greedy-like nature of the iterative optimization algorithms: when some regions of the reduction score is fixed and used as boundary conditions for updates, the Viterbi search sometimes cannot reduce notes even for smaller values of $\zeta(m)$.
Comparing iterative optimization methods in cases of target difficulty values $(30,30,40)$ and $(40,40,50)$, we find that while the {\it Iterated Gaussian} method has the largest additional-note rate, it has the least values for most difficulty evaluation measures.
If the additional-note rate increases with the fidelity to the original ensemble score, this indicates the {\it Iterated Gaussian} method has the ability to efficiently increase the fidelity while retaining low difficulty values.
This is probably because the Gaussian model is used for calculating difficulty measures.

\subsection{Subjective Evaluation}

We conduct a subjective evaluation experiment to evaluate the quality of reduction scores by the proposed algorithms\footnote{Readers who wish to have access to the raw experimental data and source code should contact the authors.}.
In particular, we examine how much of the additional notes (notes other than melodic and bass notes) are actually playable and how the musical quality such as fidelity and difficulty changes with varying target difficulty values.
For this, we asked professional piano arrangers to evaluate the piano reductions generated by the {\it Iterated Fingering}, {\it Iterated Gaussian}, and {\it Iterated Distance} methods with three sets of target difficulty values $(15,15,30)$, $(30,30,40)$, and $(40,40,50)$.
Two music arrangers participated in the evaluation and each reduction score was evaluated by one of them.
Evaluators are provided manually typeset reduction scores, the input condensed scores, and corresponding audio files of the 10 tested musical pieces, which are uploaded to the accompanying demo page\footnote{\url{http://pianoarrangement.github.io/demo.html}}.
The evaluation metrics are as follows:
\begin{itemize}
\item Musical fidelity (10 steps; $1$: not faithful at all, $\ldots$, $10$: very faithful) --- How the reduction score is faithful to the original ensemble score in terms of music acoustics.
\item Subjective difficulty (10 steps; $1$: very easy, $\ldots$, $10$: very difficult) --- How difficult the reduction score is for playing with two hands.
\item Musical naturalness (10 steps; $1$ very unnatural, $\ldots$, $10$: very natural) --- How natural the reduction score is as a piano score.
\item Number of unplayable notes $N_{\rm unp}$ --- How many notes and which notes should be removed from the reduction score to make it playable by a skillful pianist. We define the unplayable-note rate $A_{\rm unp}$, a quantity normalized by the number of additional notes:
\begin{equation}
A_{\rm unp}=\frac{N_{\rm unp}}{\#R-\#\mathcal{M}-\#\mathcal{B}}.
\end{equation}
\end{itemize}
\begin{figure*}[t]
\centering
{\includegraphics[clip,width=1.3\columnwidth]{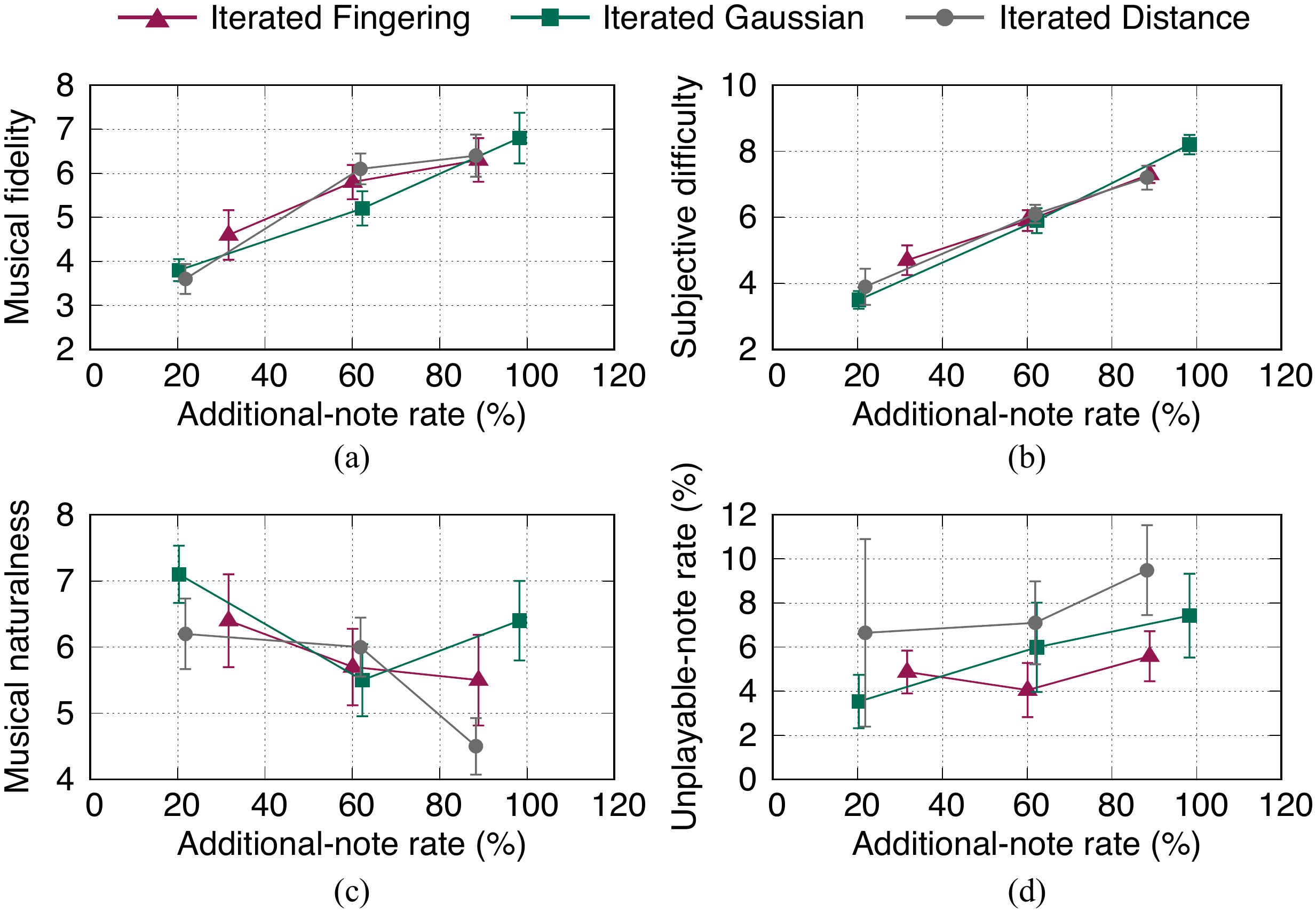}}
\caption{Subjective evaluation results. For each method, the average results for the three sets of target difficulty are indicated with points. Bars indicate their standard errors.}
\label{fig:EvalResult}
\end{figure*}
Results are summarized in Fig.~\ref{fig:EvalResult}, where statistics (mean and standard deviation) are shown for each evaluation metrics and for each method.
The results in Figs.~\ref{fig:EvalResult}(a) and \ref{fig:EvalResult}(b) indicate that subjective difficulty and musical fidelity monotonically increase with the additional note rate, which confirms the ability of the proposed methods for controlling performance difficulty.
For these two quantities, few differences can be found in the results for the three methods.
The result in Fig.~\ref{fig:EvalResult}(c) shows that musical naturalness tends to decrease when increasing the additional-note rate.
This can be understood from the fact when $A_{\rm add}\simeq0$ the reduction score consists mostly of melodic and bass notes, which should have high naturalness, and for larger $A_{\rm add}$ it becomes more demanding for the models to retain naturalness.
For the highest difficulty case with target difficulty values $(40,40,50)$ and $A_{\rm add}\sim90\%$--$100\%$, the {\it Iterated Gaussian} and {\it Iterated Fingering} methods outperform the baseline {\it Iterated Distance} method.
This suggests the importance of incorporating sequential dependence of pitches in the piano score model for improving musical naturalness.

The result in Fig.~\ref{fig:EvalResult}(d) shows that, especially in the high difficulty regime, the unplayable-note rate is reduced by incorporating sequential dependence of pitches in the piano score model and even further so by incorporating the fingering motion.
This suggests that although the same difficulty measure is used and it is not a perfect measure for describing the real difficulty of a piano score, a better piano score model can generate reduction scores with less unplayable notes.

\subsection{Example Results and Discussions}
\label{sec:ExamplesAndDiscussions}

%
\begin{figure*}[t]
\centering
{\includegraphics[clip,width=1.9\columnwidth]{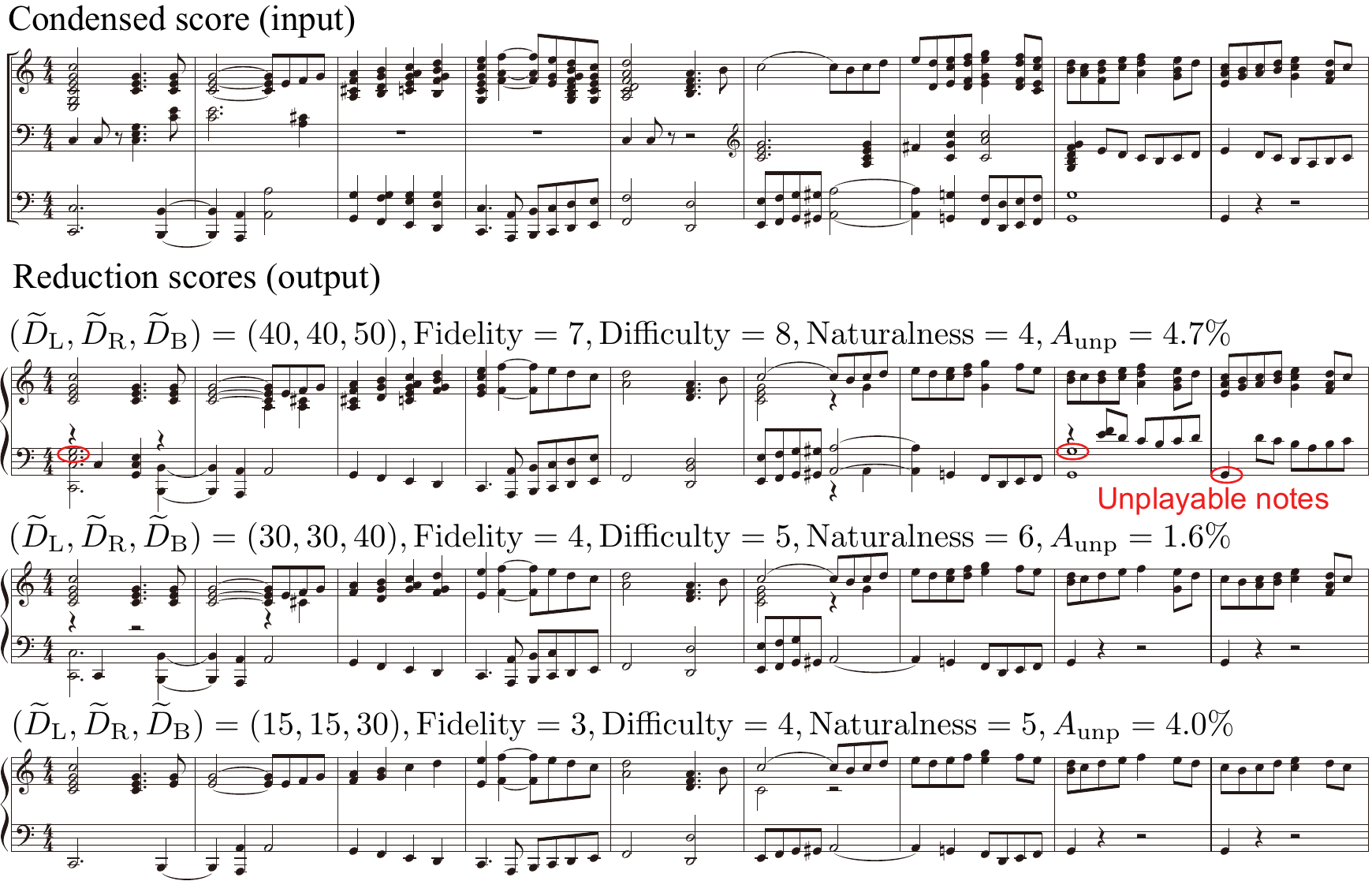}}
\caption{Examples of piano reduction scores obtained by the {\it Iterated Fingering} method (Wagner: Prelude to Die Meistersinger von N\"{u}rnberg). For clear illustration, only the first 9 bars from a 27-bar excerpt in the test data are shown. Unplayable notes indicate those identified by the evaluator.}
\label{fig:Ex}
\end{figure*}
Examples of piano reduction scores obtained by the {\it Iterated Fingering} method are shown in Fig.~\ref{fig:Ex}, together with the results of the subjective evaluation (the evaluation scores are given for the whole piece including the part not shown in the figure)\footnote{See \url{http://pianoarrangement.github.io/demo.html} for more examples with sound files.}.
We see that results for larger target difficulty values have more harmonizing notes and are given larger fidelity and subjective-difficulty values.
In the cases with $(\widetilde{D}_{\rm L},\widetilde{D}_{\rm R},\widetilde{D}_{\rm B})=(15,15,30)$ and $(30,30,40)$, all notes are playable in the shown section and the latter case has a larger musical-naturalness value.
On the other hand, there are several unplayable notes in the case with $(\widetilde{D}_{\rm L},\widetilde{D}_{\rm R},\widetilde{D}_{\rm B})=(40,40,50)$, which leads to a smaller musical-naturalness value.

We were informed from the evaluators (professional music arrangers) that keeping more notes in a reduction score does not always improve musical naturalness.
One reason is that flexibility for performance expression can be reduced by adding too many notes.
We have therefore two important directions to further improve the piano reduction methods.
One is to construct a more precise fingering model and difficulty measures based on it.
However, as we discussed in section \ref{sec:FingeringModelAndDifficulty}.\ref{sec:Difficulty}.\ref{sec:DifficultyEvaluation}, a more complex model typically requires more training data for appropriate learning.
Since a large-scale fingering dataset is currently not available, construction of such a dataset is also an important issue.
Another is to incorporate more musical knowledge in the piano reduction model, particularly on harmonic aspects (e.g.\ completion of chordal notes and voice leading) and cognitive aspects (e.g.\ restricting notes over melodic notes to avoid mishearing of melodies).

Other left issues are identification of melodic and bass notes and score typesetting for reduction scores, which are manually done currently.
As for the identification of melodic and bass notes, a simple method of taking the instrument part with the highest (lowest) mean pitch as the melody (bass) part for each bar can reproduce $40.6\%$ ($57.0\%$) of the indications in our test data.
While this calls for a more refined method for automatically estimating the melodic and bass notes, we noticed that the choice is also subjective and it may be important to leave room for user preferences.

Finally, since the evaluation is subjective, it is also important to look at multiple ratings given by different evaluators. Such a large-scale subjective evaluation would be significant for revealing finer relations between human's evaluation and the model's prediction.

\section{Conclusion}

We have described quantitative measures of performance difficulty for piano scores and a piano reduction method that can control the difficulty values based on statistical modelling.
We followed the quantification of performance difficulty using statistical models proposed in \cite{Nakamura2014} and found that the difficulty values can be used as indicators of performance errors.
For the current amount of training data, we also found that the difficulty measures based on the Gaussian model yields the best accuracy of predicting performance errors.
The problem of piano reduction is formalized as a statistical optimization problem following the framework of \cite{Nakamura2015}, and we improved the optimization method by proposing an iterative method.
We confirmed the efficacy of the iterative optimization method and the algorithms are shown to be able to control subjective difficulty and musical fidelity.
It was also found that incorporating sequential dependence and fingering motion in the piano-score model by using the Gaussian and fingering model improves generated reduction scores in terms of musical naturalness and the rate of unplayable notes.
Directions for further improvements were also discussed.

Whereas it has been assumed that the same difficulty measures apply universally for all players in this study, they can be different for individual players depending on, for example, the size of hands.
In the present framework, part of such individuality can be expressed by adapting the fingering model to individual players.
This model adaptation can be realized in principle if one has a sufficient amount of musical scores that have been already played by an individual player.
Another interesting direction is to adapt an individual's fingering model using the frequency of errors in his/her performance data, which could reduce the amount of necessary data.

A limitation of the present model is that timing errors and other rhythmic aspects are not considered.
Rhythmic features may become important especially in polyrhythmic passages in which the left and right hand parts have contrasting rhythms (e.g.\ two against three rhythms).
In such cases, the sum of difficulty values for the two hands may underestimate the total difficulty.
To properly deal with these problems, it would be necessary to incorporate a performance timing model and interdependence between the two hands into the present framework.

The present formulation of combining a musical-score model and an edit model can also be applied to other forms of music arrangement if one replaces the piano fingering model with an appropriate score model of the target instrumentation/style and adapt the edit model for relevant edit operations.
For example, if we combine a score model for jazz music and a proper edit model, it would be possible to develop a method for arranging a given piece in the rock music style (or other styles) into a piece in the jazz style.

Although this study has focused on piano arrangement, the framework can also be useful for music transcription \cite{Benetos2013}.
In music transcription, musical-score models play an important role to induce an output score to be an appropriate one that respects musical grammar, style of target music, etc.\ \cite{Raczynski2013b,Ycart2018}.
Especially in piano transcription, results of multi-pitch detection contain a significant amount of spurious notes (false positives), which often make the transcription results unplayable \cite{Nakamura2018}.
By integrating the present piano-score model and an acoustic model (instead of the edit model) and applying the method for optimization developed in this study, one can impose constraints on performance difficulty of transcription results and reduce these spurious notes.

%

\section*{Financial Support}

This study was partially supported by JSPS KAKENHI Nos.\ 26700020, 16H01744, 16J05486, 16H02917, and 16K00501, and JST ACCEL No.\ JPMJAC1602.

\section*{Statement of interest}

None.

\vskip2pc

%
%

%


\vskip2pc

\noindent \large \textbf{Biographies}

\vskip2pc

\noindent\normalsize\textbf{Eita Nakamura} received his Ph.D.\ degree in physics from the University of Tokyo, Tokyo, Japan, in 2012. After having been a Postdoctoral Researcher at the National Institute of Informatics, Meiji University, and Kyoto University, Kyoto, Japan, he is currently a Research Fellow of Japan Society for the Promotion of Science. His research interests include music modeling and analysis, music information processing, and statistical machine learning.

\vskip2pc

\noindent\textbf{Kazuyoshi Yoshii} received his M.S.\ and Ph.D.\ degrees in informatics from Kyoto University, Kyoto, Japan, in 2005 and 2008, respectively. He is currently a Senior Lecturer at the Graduate School of Informatics, Kyoto University, and concurrently the Leader of the Sound Scene Understanding Team, RIKEN Center for Advanced Intelligence Project, Tokyo, Japan. His research interests include music analysis, audio signal processing, and machine learning.

\vskip2pc


\begin{thebibliography}{99}
%
\bibitem{Chiu2009}
Chiu, S.-C.; Shan, M.-K.; Huang, J.-L.:
Automatic system for the arrangement of piano reduction,
in {\it IEEE International Symposium on Multimedia}, San Diego, California, 2009, 459--464.
%
\bibitem{Onuma2010}
Onuma, S.; Hamanaka, M.:
Piano arrangement system based on composers' arrangement processes,
in {\it International Computer Music Conference}, New York, 2010, 191--194.
%
\bibitem{Huang2012}
Huang, J.-L.; Chiu, S.-C.; Shan, M.-K.:
Towards an automatic music arrangement framework using score reduction.
ACM Transactions on Multimedia Computing, Communications, and Applications, {\bf8(1)} (2012), 8:1--8:23.
%
\bibitem{Nakamura2015}
Nakamura, E.; Sagayama S.:
Automatic piano reduction from ensemble scores based on merged-output hidden Markov model,
in {\it International Computer Music Conference}, Denton, Texas, 2015, 298--305.
%
\bibitem{Takamori2017}
Takamori, H.; Sato, H.; Nakatsuka, T.; Morishima, S.:
Automatic arranging musical score for piano using important musical elements,
in {\it International Sound and Music Computing Conference}, Aalto, 2017, 35--41.
%
\bibitem{Tuohy2005}
Tuohy, D.R.; Potter, W.D.:
A genetic algorithm for the automatic generation of playable guitar tablature,
in {\it International Computer Music Conference}, Barcelona, 2005, 499--502.
%
\bibitem{Hori2012}
Hori, G.; Yoshinaga, Y.; Fukayama, S.; Kameoka, H.; Sagayama, S:
Automatic arrangement for guitars using hidden Markov model,
in {\it International Sound and Music Computing Conference}, Copenhagen, 2012, 450--456.
%
\bibitem{Hori2013}
Hori, G.; Kameoka, H.; Sagayama, S:
Input-output HMM applied to automatic arrangement for guitars.
J.\ Info.\ Processing Soc.\ Japan, {\bf21(2)} (2013), 264--271.
%
\bibitem{Maekawa2006}
Maekawa, H.; Emura, N.; Miura, M.; Yanagida, M.:
On machine arrangement for smaller wind-orchestras based on scores for standard wind-orchestras,
in {\it International Conference on Music Perception and Cognition}, Bologna, 2006, 268--273.
%
\bibitem{Crestel2017}
Crestel, L.; Esling, P.:
Live orchestral piano, a system for real-time orchestral music generation,
in {\it International Sound and Music Computing Conference}, Espoo, 2017, 434--442.
%
\bibitem{Chiu2012}
Chiu, S.-C.; Chen, M.-S.:
A study on difficulty level recognition of piano sheet music,
in {\it IEEE International Symposium on Multimedia}, Irvine, California, 2012, 17--23.
%
\bibitem{Sebastien2012}
S\'ebastien, V.; Ralambondrainy, H.; S\'ebastien, O.; Conruyt, N.:
Score analyzer: Automatically determining scores difficulty level for instrumental e-learning,
in {\it International Conference on Music Information Retrieval}, Porto, 2012, 571--576.
%
\bibitem{StatisticalTranslation}
Brown, P.F.; Pietra, V.J.D.; Pietra, S.A.D.; Mercer, R.L.:
The mathematics of statistical machine translation: parameter estimation.
Computational Linguistics, {\bf 19(2)} (1993), 263--311.
%
\bibitem{Parncutt1997}
Parncutt, R.; Sloboda, J.A.; Clarke, E.F.; Raekallio, M.; Desain, P.:
An ergonomic model of keyboard fingering for melodic fragments.
Music Perception, {\bf14(4)} (1997), 341--382.
%
\bibitem{Hart2000}
Hart, M.; Tsai, E.:
Finding optimal piano fingerings.
The UMAP Journal, {\bf21(1)} (2000), 167--177.
%
\bibitem{Kasimi2007}
Al Kasimi; A., Nichols, E.; Raphael, C:
A simple algorithm for automatic generation of polyphonic piano fingerings,
in {\it International Conference on Music Information Retrieval}, Vienna, 2007, 355--356.
%
\bibitem{Yonebayashi2007}
Yonebayashi, Y.; Kameoka, H.; Sagayama, S.:
Automatic decision of piano fingering based on a hidden Markov models,
in {\it International Joint Conference on Artificial Intelligence}, Hyderabad, 2007, 2915-2921.
%
\bibitem{Nakamura2014}
Nakamura, E.; Ono, N.; Sagayama S.:
Merged-output HMM for piano fingering of both hands,
in {\it International Conference on Music Information Retrieval}, Taipei, 2014, 531--536.
%
\bibitem{Rabiner1989}
Rabiner, L.:
A tutorial on hidden Markov models and selected applications in speech recognition.
Proc.\ IEEE, {\bf77(2)} (1989), 257--286.
%
\bibitem{Nakamura2017TASLP}
Nakamura, E.; Yoshii, K.; Sagayama, S.:
Rhythm transcription of polyphonic piano music based on merged-output HMM for multiple voices.
IEEE/ACM Trans.\ on Audio, Speech and Language Processing, {\bf25(4)} (2017), 794--806.
%
\bibitem{Nakamura2017}
Nakamura, E.; Yoshii, K.; Katayose, H.:
Performance error detection and post-processing for fast and accurate symbolic music alignment,
in {\it International Conference on Music Information Retrieval}, Suzhou, 2017, 347--353.
%
\bibitem{Benetos2013}
Benetos, E.; Dixon, S.; Giannoulis, D.; Kirchhoff, H.; Klapuri, A.:
Automatic music transcription: Challenges and future directions.
J.\ Intelligent Information Systems, {\bf41(3)} (2013), 407--434.
%
\bibitem{Raczynski2013b}
Raczy\'{n}ski, S.; Vincent, E.; Sagayama, S.:
Dynamic Bayesian networks for symbolic polyphonic pitch modeling.
IEEE Trans.\ on Audio, Speech, and Language Processing, {\bf21(9)} (2013), 1830--1840.
%
\bibitem{Ycart2018}
Ycart, A.; Benetos, E.:
Polyphonic music sequence transduction with meter-constrained LSTM networks,
in {\it IEEE International Conference on Acoustics, Speech, and Signal Processing}, Calgary, 2018, 386--390.
%
\bibitem{Nakamura2018}
Nakamura, E.; Benetos, E.; Yoshii, K.; Dixon, S.:
Towards complete polyphonic music transcription: Integrating multi-pitch detection and rhythm quantization,
in {\it IEEE International Conference on Acoustics, Speech, and Signal Processing}, Calgary, 2018, 101--105.
%
\end{thebibliography}
\end{document}